%% file: main.tex
\documentclass[10pt,a4paper,twocolumn]{article}

\usepackage[
  a4paper,
  top=1.80cm,
  bottom=2.05cm,
  left=1.70cm,
  right=1.70cm,
  columnsep=0.70cm,
  headheight=13pt,
  footskip=0.85cm
]{geometry}
\usepackage{fontspec}
\usepackage{microtype}
\usepackage{amsmath,amssymb,bm}
\usepackage{booktabs}
\usepackage{makecell}
\usepackage{multirow}
\usepackage{tabularx}
\usepackage{array}
\usepackage{enumitem}
\usepackage[section]{placeins}
\usepackage{graphicx}
\usepackage{titlesec}
\usepackage{indentfirst}
\usepackage{stfloats}
\usepackage{flushend}
\usepackage[font=footnotesize,labelfont=bf,labelsep=period,skip=4pt]{caption}
\usepackage{url}
\usepackage{xspace}
\usepackage[numbers,sort&compress]{natbib}
\usepackage{xcolor}

\definecolor{cvprblue}{rgb}{0.12,0.36,0.62}
\usepackage[unicode,breaklinks,colorlinks,allcolors=cvprblue]{hyperref}
\hypersetup{
  pdftitle={Hi-FLoop: Hierarchical State-Feedback Loops for Multi-Timescale World Modeling},
  pdfauthor={Rx Fan and Z Han},
  pdfsubject={Multi-agent closed-loop traffic simulation},
  pdfkeywords={traffic simulation, world model, joint future branches, state feedback}
}

\renewcommand{\arraystretch}{1.05}

\allowdisplaybreaks[2]
\titleformat{\section}{\large\bfseries}{\thesection}{0.55em}{}
\titlespacing*{\section}{0pt}{2.0ex plus .5ex minus .2ex}{0.8ex}
\titleformat{\subsection}{\normalsize\bfseries}{\thesubsection}{0.55em}{}
\titlespacing*{\subsection}{0pt}{1.6ex plus .4ex minus .2ex}{0.55ex}
\titleformat{\paragraph}[runin]{\normalsize\bfseries}{}{0pt}{}[.]
\titlespacing*{\paragraph}{0pt}{1.2ex plus .3ex minus .2ex}{0.55em}
\setlist[itemize]{leftmargin=1.45em,labelsep=.45em,itemsep=1pt,topsep=2pt,parsep=0pt,partopsep=0pt}
\newcommand{\model}{\textsc{Hi-FLoop}\xspace}
\newcolumntype{Y}{>{\raggedright\arraybackslash}X}
\newcolumntype{P}[1]{>{\raggedright\arraybackslash}p{#1}}

\title{Hi-FLoop: Hierarchical State-Feedback Loops\\[-0.08em]
for Multi-Timescale World Modeling}
\author{Rx Fan \qquad Z Han\thanks{Corresponding author.}\\[0.18em]
\normalsize School of Systems Science, Beijing Normal University\\[-0.02em]
\small \texttt{Gyro@mail.bnu.edu.cn} \quad \texttt{zhan@bnu.edu.cn}}
\date{}

\makeatletter
\renewcommand{\@maketitle}{%
  \newpage
  \null
  \vspace*{-1.05cm}%
  \begin{center}%
    {\LARGE\bfseries\@title\par}%
    \vspace{0.72em}%
    {\large\lineskip .45em\begin{tabular}[t]{c}\@author\end{tabular}\par}%
  \end{center}%
  \vspace{0.65em}%
}
\makeatother

\begin{document}
\maketitle

\input{sections/frontmatter_related}
\input{sections/method}
\input{sections/experiments_conclusion}

{\footnotesize
\bibliographystyle{ieeenat_fullname}
\bibliography{refs}
}

\end{document}

%% file: sections/frontmatter_related.tex
\begin{abstract}
\setlength{\parindent}{2em}
Multi-agent traffic simulation seeks diverse, coordinated, and physically realistic futures from maps and observed history. Long-horizon closed-loop generation must reconcile multiple decision time scales while its context evolves with generated states. Existing methods often unfold long futures from an initial scene and resolve intent, interaction, and motion monolithically, weakening cross-scale consistency and adaptation. Multimodal rollout poses a further consistency problem: independently reselecting modes across agents or commits can stitch together incompatible futures instead of preserving a coherent joint branch. We present \model{}, a branch-consistent multi-timescale state-feedback framework. Eight scene-level Worlds represent joint hypotheses; all agents share one selected World identity throughout all 16 commits of an 8-second rollout, while Goal, Preview, and Control states adapt within that branch. An 8-second Goal anchors intent, a 2-second Preview coordinates interactions, and 1-second Control produces physical motion. Every 0.5-second commit feeds back only its executed prefix as new facts, while unexecuted hypotheses never enter factual memory. Joint Preview Interaction induces a sparse directed future graph and uses conflict probabilities and signed arrival-time differences to refine interaction-aware motion. For generated-state recovery, a prefix-frozen A$\rightarrow$B cascade transfers typed physical state and the branch index---but no latent state---from a frozen prefix model to an independently parameterized recovery model. On the full H-D public-validation split of 955 scenarios, the S2.1 cascade obtains an 8-second scene-joint ADE-at-joint-minFDE@8/joint-minFDE@8 of $2.048/6.384$~m when one World must explain all evaluated agents. Agent-centric oracle-minADE@8 is $0.526$~m at 6 seconds and $0.875$~m at 8 seconds.
\end{abstract}

\section{Introduction}

Open-loop motion prediction infers future motion from a fixed observed history. Closed-loop traffic simulation instead feeds model outputs back into the system, so the prediction problem itself changes with the generated state. Errors in position, velocity, or heading during the first few steps not only affect an individual agent's subsequent trajectory, but may also alter agent neighborhoods, map relations, conflict locations, and right-of-way order, thereby changing the appropriate responses of other traffic participants. Closed-loop error therefore accumulates over time and propagates through scene interactions. In heterogeneous scenes containing vehicles, pedestrians, and cyclists, a model must additionally reconcile long-term behavioral direction, local interaction coordination, and type-specific motion feasibility at different time scales; marginal trajectories that appear plausible in isolation may still conflict when composed. Long-horizon simulation should consequently be assessed not only by single-trajectory errors, but also by joint interaction consistency, safety, map and kinematic compliance, and coverage of the generated distribution~\cite{montali2023wosac}.

Multi-agent motion predictors now provide strong factual encoding, intent conditioning, and local trajectory refinement. QCNet models scene relations with query-centric representations~\cite{zhou2023qcnet}, while MTR and MTR++ organize multimodal prediction and global-to-local trajectory refinement around motion-intention queries~\cite{shi2022mtr,shi2023mtrpp}. For predictors conditioned on a fixed history, however, naively invoking the model in a rolling fashion does not by itself define state read/write semantics across commits. Meanwhile, traffic simulators such as TrafficSim, TrafficBots, and BehaviorGPT have introduced generated-state feedback or reactive rollout~\cite{suo2021trafficsim,zhang2023trafficbots,zhou2024behaviorgpt}. Building on these advances, we focus on explicit causal interfaces among long-term joint intent, rolling interaction plans, and physical execution: how a shared joint branch should remain consistent across both agents and commits, which local plans may be revised as new facts arrive, how continuous actions are grounded into executable states, and what information is eligible to enter the factual memory. Without these boundaries, short-term avoidance may override long-term direction, per-agent mode choices may break scene-level branch consistency, and unexecuted futures may be conflated with observed facts. Section~\ref{sec:related} provides a more detailed comparison of tasks and methods.

We formulate this problem as \emph{persistent multi-timescale joint world modeling}. Long-term joint intent, mid-term interaction plans, and short-term executable motion are not merely different output heads attached to a shared latent state; they are generative states with distinct predictive horizons, temporal persistence, and adaptation scopes. The long-term layer maintains the joint future hypothesis underlying an entire rollout, the mid-term layer continually revises local coordination among agents according to executed facts, and the short-term layer converts the current plan into physical motion. After each short prefix is executed, the model constructs the next generation condition from the updated scene facts. Unexecuted futures always remain planning hypotheses and are never mixed with physical events that have already occurred. Long-horizon closed-loop modeling must therefore jointly address cross-scale decision consistency, consistency of a shared joint branch across agents and commits, and dynamic adaptation of interactions to new states.

Under this formulation, \model{} adopts a QCNet-style query-centric factual encoder and draws on the intent-conditioning and local-refinement paradigms of the MTR family. These mature components form the implementation basis rather than our novelty claim. On top of them, eight scene-level Worlds instantiate joint hypotheses. Once a World is selected, its discrete identity is shared by every agent and held fixed across all commits, while its continuous Goal and Preview states remain adaptive to newly executed facts. A mid-term Preview instantiates long-term intent as a local route and passing order. Across two rounds of interaction refinement, the model reconstructs a sparse relation topology from the updated joint Preview and updates continuous conflict and timing attributes; local Control then translates the interaction-conditioned plan into executable physical states. For generated-state recovery, a frozen model produces a closed-loop prefix and an independently parameterized model learns its continuation; routing remains index-consistent across the boundary, while the recovery model reconstructs its own latent representation from transferred physical facts. In this way, explicit causal boundaries unify long-term joint hypotheses, rolling interaction adaptation, short-term physical execution, and generated-state training.

Our main contributions are as follows:
\begin{itemize}
  \item \textbf{A multi-timescale state decomposition with executed-state feedback.} We assign long-term behavioral intent, mid-term interactions, and short-term motion evolution distinct predictive horizons, lifetimes, and causal scopes, and close the recurrence through executed-state feedback. This formulation reconciles cross-scale decision consistency with continual adaptation to generated-state evolution within a unified generation process.
  \item \textbf{Scene-level joint-branch consistency across agents and commits.} We instantiate scene-level World branches whose selected discrete identity is shared by all agents and retained throughout the rollout, while continuous Goal and Preview states adapt within that branch as executed facts evolve. This separates the persistence of a joint future hypothesis from the adaptivity required for rolling generation.
  \item \textbf{Preview-driven alternating structure--attribute interaction reasoning.} Over a rolling Preview, we alternate between reconstructing sparse directed relation topology and updating continuous conflict and timing attributes. Plan changes can thus revise the interaction structure, while interaction reasoning in turn refines the plan, making mid-term coordination an explicit condition for short-term motion generation.
  \item \textbf{Prefix-frozen recovery on generated-state distributions.} A frozen prefix model induces the model-generated states on which an independent recovery model learns the continuation. This exposes recovery training to deployment-like inputs while preventing suffix optimization from interfering with the learned prefix generator.
\end{itemize}
On the full H-D public-validation split of 955 scenarios, the evaluated S2.1 A$\rightarrow$B cascade attains an 8-second scenario-joint ADE-at-joint-minFDE@8/joint-minFDE@8 of $2.048/6.384$~m. Independent agent-centric ADE selection yields oracle-minADE@8 of $0.526$~m at 6 seconds and $0.875$~m at 8 seconds. Section~\ref{sec:method} details the 8-/2-/1-second horizons, 0.5-second commits, World indexing, JPI refinement, and generated-state recovery.

\section{Related Work}
\label{sec:related}

\subsection{Multi-Agent Motion Prediction and Closed-Loop Traffic Generation}

Multi-agent motion prediction has progressed from per-agent marginal trajectories to multimodal joint futures. AgentFormer and Scene Transformer jointly model social--temporal relations or multi-agent dependencies within a scene~\cite{yuan2021agentformer,ngiam2022scenetransformer}. M2I and FJMP factor interactive prediction through directed relations, whereas JFP directly learns mutually consistent joint futures for multiple agents~\cite{sun2022m2i,rowe2023fjmp,luo2023jfp}. MotionLM and MotionDiffuser represent joint uncertainty using discrete motion tokens and diffusion distributions, respectively~\cite{seff2023motionlm,jiang2023motiondiffuser}. Query-conditioned predictors are most closely related to our architecture: QCNet encodes scene relations with query-centric representations, MTR/MTR++ use intention queries for global target localization and local motion refinement, and BiFF further fuses multi-level future information~\cite{zhou2023qcnet,shi2022mtr,shi2023mtrpp,zhu2023biff}. We adopt QCNet's query-centric factual encoding and the intent-conditioning and local-refinement paradigms of the MTR family; query representations, multimodal goals, and hierarchical decoding are not themselves our contributions. These prediction methods primarily address \emph{what futures are possible} given a fixed history, whereas state read/write semantics and continual replanning across multiple execution commits pose a distinct problem.

Data-driven traffic generation additionally requires multiple agents to respond continually on states produced by the model itself. TrafficSim and TrafficBots learn scene-level stochasticity and reactive driving policies, respectively; BehaviorGPT generates behavior autoregressively in next-patch form, while Trajeglish and SMART cast traffic generation as next-token modeling~\cite{suo2021trafficsim,zhang2023trafficbots,zhou2024behaviorgpt,philion2024trajeglish,wu2024smart}. Among generative approaches, CTG produces controllable trajectories with guided diffusion, SceneDiffuser supports constraint-aware initialization and closed-loop rollout, and CCDiff further introduces causal compositional reasoning~\cite{zhong2023ctg,jiang2024scenediffuser,lin2025ccdiff}. SceneDiffuser++ scales the generation of scenes, dynamic agents, and traffic signals to the city level, while ProSim controls closed-loop multi-agent rollouts through numerical, categorical, or textual prompts~\cite{tan2025scenediffuserpp,tan2025prosim}. Rather than reintroducing hierarchical structure, interaction modeling, or generated-state feedback, we define a unified causal contract for these components across commits: the selected World index and its associated query define a persistent joint branch for all agents throughout a rollout, the Goal persists as a slow state across commits, the Preview updates local routes and passing order from new facts, and Control writes only its executed prefix back to the factual memory after dynamics propagation. These boundaries among a joint hypothesis, a revisable plan, and physical facts are orthogonal to whether the underlying model uses queries, tokens, or diffusion.

\paragraph{Distinction from Hierarchical Closed-Loop Simulators.}
Hierarchical planning and closed-loop feedback both have clear precedents. BITS separates high-level intent inference from low-level driving behavior, while MixSim and HMSim adopt hierarchical traffic-simulation frameworks~\cite{xu2023bits,suo2023mixsim,liu2026hmsim}. Beyond Self-Play conditions low-level continuous motion on high-level multi-agent interaction reasoning and adapts to closed-loop deployment through recovery supervision~\cite{zhang2026beyondselfplay}. These methods demonstrate the benefit of policy--motion hierarchies for closed-loop behavior, but do not specify how a scene-level multimodal joint branch retains its identity across agents and repeated commits, nor do they distinguish the lifecycles of a long-term Goal, a mid-term interaction Preview, short-term Control, and executed facts. This set of cross-commit state semantics is precisely the focus of \model{}: World identity persists, Goals and interactions adapt to new facts within the branch, and only the prefix executed by the dynamics can be written back to history.

For longer horizons, TrafficBots, ProSim, and AutoWorld advance traffic simulation through reactive policies, promptable closed-loop generation, and self-supervised world modeling, respectively~\cite{zhang2023trafficbots,tan2025prosim,pourkeshavarz2026autoworld}. RosettaSim organizes scene topology, agent states, and the introduction of new agents into variable-length structured autoregressive sequences, and evaluates long-horizon simulation with retrieval-based references~\cite{xiao2026rosettasim}. Its central concerns are long-sequence representation, dynamic agent cardinality, and long-horizon evaluation. In contrast, \model{} targets persistent causal evolution within a fixed joint World: the same World retains its routing identity across all agents and 16 commits; the long-term Goal, rolling interaction Preview, and short-term Control evolve according to their respective state semantics; and unexecuted plans remain isolated from executed facts. Our distinction is therefore not generic long-sequence autoregression, but joint branch consistency across agents and rolling time together with a causal contract for multi-timescale state feedback.

\subsection{Training and Post-Training for Closed-Loop Traffic Generation}

Closed-loop training can be characterized along three separable dimensions: the learning signal, the source of states, and the gradient scope. BITS decomposes high-level intent inference and low-level driving behavior through hierarchical imitation, while SMART supervises multi-agent motion generation with next-token prediction~\cite{xu2023bits,wu2024smart}. To address the covariate shift between log-state training and the states encountered in model rollouts, LASIL constructs learner-aware supervision examples from the learner's state distribution. CAT-K instead selects, from the policy's top-$K$ action tokens, the token whose resulting next state is closest to the ground truth and continues closed-loop fine-tuning from that state~\cite{guo2024lasil,zhang2025catk}. TrafficSim and ProSim perform differentiable closed-loop rollout with backpropagation through time (BPTT) on generated states~\cite{suo2021trafficsim,tan2025prosim}. TrafficBots also rolls out autoregressively, but stops gradients through the action path while retaining cross-time gradients through the state path~\cite{zhang2023trafficbots}. Rectify, Don't Regret detaches the computation graph between adjacent replanning steps to prevent future supervision from creating a non-causal gradient shortcut through induced states~\cite{yadav2026rectify}. Classical truncated BPTT (TBPTT) reduces the computation and memory costs of long sequences with a finite backward window, at the cost of truncating credit assignment across windows~\cite{williams1990tbptt}.

Rectify focuses on gradient paths when the same predictor is trained against receding target trajectories, while surrounding agents continue to follow log replay. Our S2.1 training extension instead assigns the learned prefix and generated-state recovery to disjointly parameterized Models~A and B, and transfers only observable states, admissible exogenous context, and discrete routing labels. Suffix gradients therefore cannot alter the prefix model at the parameter level. The methodological claim lies in jointly isolating temporal responsibility, parameter ownership, and the causal state interface, rather than in detachment or TBPTT itself; the branch-consistent multi-timescale generation framework of \model{} likewise does not depend on this gradient operation for its definition.

Reinforcement learning and post-training can complement supervised objectives with closed-loop signals for collision avoidance, compliance, behavioral quality, or diversity. Shiroshita et al.\ select policy sets subject to driving-competence constraints and use intrinsic rewards to diversify behavior across policies, evaluating behavioral coverage with trajectory-based metrics~\cite{shiroshita2020diverse}. RL fine-tuning and TrafficRLHF refine traffic behavior models using closed-loop rewards and human preferences, respectively~\cite{peng2024rlfinetune,cao2024trafficrlhf}. ForSim introduces step-wise forward simulation into group-relative policy fine-tuning: at each virtual step, the focal traffic agent propagates candidate modes that are spatiotemporally aligned with a reference trajectory, while other agents repeatedly repredict from the updated state, balancing within-mode continuity, physical feasibility, and interactive responsiveness~\cite{chen2026forsim}. The related Plan-R1 aligns ego planning for safety and feasibility using rule-based rewards and VD-GRPO after expert-trajectory pretraining, whereas SMART-R1 applies R1-style reinforcement fine-tuning to multi-agent token-based traffic simulation~\cite{tang2026planr1,pei2026smartr1}. These approaches emphasize reward alignment and post-training rollouts; our focus is the joint-branch and multi-timescale state interface in supervised closed-loop generation, making the two directions orthogonal at the level of optimization objectives.

%% file: sections/method.tex
\section{Method}
\label{sec:method}

\subsection{Overall Architecture and Closed-Loop State}

\input{figures/hi_floop_architecture}

Figure~\ref{fig:architecture} provides an overview of the proposed framework. Given a scene with $N$ vehicles, pedestrians, and cyclists, the input history $\bm H_0$ contains 11 frames of agent states sampled at 10\,Hz, while the map $\bm M$ contains polylines, boundaries, topology, and available semantics. The objective is to jointly generate 80 future frames per agent over an 8-second horizon. A complete rollout is divided into $C=16$ commits. At each commit, the model generates one second, or 10 control steps, but executes only the first $K_e=5$ steps (0.5 seconds). This plan-longer-than-commit design preserves a local interaction look-ahead while allowing the model to condition on newly generated states every 0.5 seconds.

Table~\ref{tab:state-contract} defines the semantics of the hierarchy. Slow, Medium, and Fast refer to prediction horizons, state persistence, and write-back permissions rather than three independently scheduled inference rates; all three levels are accessed at every 0.5-second commit. Control serves as the action interface from a plan to physical facts, rather than a fourth persistent semantic state.

\begin{table*}[t]
\centering
\caption{Closed-loop read--write contract for the three semantic states. $\mathcal G_c$ denotes the future interaction graph at the current commit; the persistent branch state $\bm B_c$ is isolated from the factual history $\bm H_c$.}
\label{tab:state-contract}
\footnotesize
\begin{tabularx}{\textwidth}{P{3.0cm}P{3.7cm}P{4.1cm}Y}
\toprule
Semantic state & Carrier and horizon & Lifecycle and update & Write-back permission\\
\midrule
Long-horizon branch intent & $(w,\widetilde{\bm g}_c,\bm h_c^w)$; 8 seconds & $w$ remains fixed throughout the rollout; the Goal and hidden state receive bounded updates at each commit & Written only to the branch state $\bm B_c$, never to the factual encoder\\
Rolling interaction plan & $(\bm P_c,\mathcal G_c,\overline{\bm P}_c,\overline{\bm r}_c)$; 2 seconds & Replanned and warm-shifted every 0.5 seconds & Stores only the current plan and its branch summary\\
\makecell[l]{Executed\\physical facts} & $(\bm x_c,\bm H_c)$; 10\,Hz & Commits only the first five steps of the one-second Control & The only state allowed to enter $\bm H_{c+1}$ and be re-encoded\\
\bottomrule
\end{tabularx}
\end{table*}

At runtime, these semantic states are maintained in two isolated containers. The factual state $\bm H_c$ contains only physical states that have been observed or executed up to commit $c$. The persistent branch state
\begin{equation}
 \begin{aligned}
 \bm B_c=(&w^*,\bm q_{w^*},\bm k_{w^*},
 \widetilde{\bm g}_{c-1},\bm h^{w^*}_c,\\
 &\overline{\bm P}_{c-1},\overline{\bm r}_{c-1},\bm\chi_c),\\[-0.2em]
 &\bm k_w=(k_{w,i})_{i=1}^{N}
 \end{aligned}
 \label{eq:branch-state}
\end{equation}
stores the fixed World identity $w^*$, its World Query $\bm q_{w^*}$ and per-agent Anchor indices $\bm k_{w^*}$, together with the slow Goal Region, the World hidden state, the previous Preview warm start, an interaction summary, and the committed quality prefix $\bm\chi_c$. Although $\bm B_c$ persists across commits, it is never treated as an observation by the factual encoder. Let $E$, $G$, $P$, $C$, and $\Phi$ denote factual encoding, slow-goal updating, Preview-based interaction planning, control decoding, and joint dynamics, respectively. One commit then follows
\begin{align}
 \bm F_c &= E(\bm H_c,\bm M), \nonumber\\
 \widetilde{\bm g}_c &=G(\bm F_c,\bm B_c), \nonumber\\
 (\bm P_c,\mathcal G_c,\bm h^w_{c+1})
 &=P(\bm F_c,\widetilde{\bm g}_c,\bm B_c), \nonumber\\
 \bm U_c&\sim C(\bm F_c,\widetilde{\bm g}_c,\bm P_c,\bm h^w_{c+1}),
 \nonumber\\
 \bm X_c^{1:10}&=\Phi(\bm x_c,\bm U_c),\nonumber\\
 \bm H_{c+1}&=\operatorname{Roll}(\bm H_c,\bm X_c^{1:5}).
 \label{eq:closed-loop-recursion}
\end{align}
Consequently, $\bm P_c$ and the unexecuted states $\bm X_c^{6:10}$ remain planning hypotheses; only $\bm X_c^{1:5}$ may enter $\bm H_{c+1}$. State feedback neither reselects the World nor changes the Goal Anchor. Instead, it adapts the continuous evolution within a persistent branch identity according to newly established facts. This distinction gives the hierarchy its causal meaning, beyond merely stacking modules with different prediction horizons.

Agent sets serving different purposes are also kept explicit. The existence mask determines which agents participate in physical updates, the output mask determines which agents require generation, the supervision mask selects entries with a valid ground-truth prefix, and the official mask is used only for final evaluation. Context agents still participate in factual encoding, interaction graphs, and safety computation. A trajectory that terminates early in future ground truth merely shortens the supervised prefix; it is neither padded as a stationary trajectory nor interpreted as a motion mode. During training, the model constructs eight lightweight World plans but recursively executes only one physical World per scene. For the official H-D export, each $w$ is combined with four control-noise streams that remain temporally correlated and persistent across commits, yielding 32 independently executed, scene-level joint rollouts. Within each rollout, all agents share the same $w$ and sample stream while maintaining separate dynamic histories; per-agent, per-commit, or post-hoc stitching across Worlds is prohibited. The four samples within a World represent within-branch execution stochasticity rather than four distinct semantic Worlds.

\subsection{Factual Scene Encoder with Causal Inputs Only}

The factual encoder adopts a QCNet-style factorized relation model. It first applies Temporal Self-Attention to the 11-frame history of each agent, retaining both sequence memory and a history summary. A polyline encoder then represents lanes, road boundaries, drivable regions, and map topology, followed by dynamic Agent--Map Attention and directed Agent--Agent relation attention. Relation edges are computed in the query-centric coordinate frame of the receiving agent and augmented with directed type-pair embeddings to distinguish, for example, vehicle$\rightarrow$pedestrian, pedestrian$\rightarrow$vehicle, and vehicle$\rightarrow$vehicle interactions. Rather than producing a single scene vector, the encoder returns the multi-source factual memory
\begin{equation}
 \mathcal{F}_{c}=\{\bm{H}^{\mathrm{hist}}_c,\bm{H}^{\mathrm{map}},
 \bm{H}^{a2m}_c,\bm{H}^{a2a}_c,\bm{H}^{\mathrm{sig}}_c\}.
\end{equation}
For the $\ell$-th layer, the map and agent relation updates can be summarized as
\begin{align}
 \bm h_{i,c}^{a2m,\ell}
 &=\operatorname{Attn}_{a2m}\!\left(
 \bm H_c^{\mathrm{map}},\bm h_{i,c}^{a2a,\ell-1},
 \{\bm e_{p\rightarrow i,c}^{m}\}\right),\nonumber\\
 \bm h_{i,c}^{a2a,\ell}
 &=\operatorname{Attn}_{a2a}\!\left(
 \{\bm h_{j,c}^{a2m,\ell}\},\bm h_{i,c}^{a2m,\ell},
 \{\bm e_{j\rightarrow i,c}^{a},t_{j\rightarrow i}\}\right),\nonumber\\
 \bm f_{i,c}
 &=\phi_f\!\left[
 \bm h_{i,c}^{\mathrm{hist}},
 \bm h_{i,c}^{a2m},
 \bm h_{i,c}^{a2a}\right],
 \label{eq:factual-relations}
\end{align}
where $t_{j\rightarrow i}$ encodes the directed agent-type pair. Keeping the memory sources separate allows the downstream Goal, Preview, and Control modules to access history, map, agent, and signal facts as needed, without relying on a single compressed scene vector.

Static map representations may be cached within a continuous rollout. After each commit, the model appends five generated states to the rolling history, retains the most recent 11 frames, and recomputes dynamic relations. During TBPTT training, static representations may be recomputed across chunks to instantiate a new computation graph, but their numerical semantics do not change with the generated future. Different Worlds may share only the static map and the initial read-only observations; they cannot share dynamic memory after their trajectories diverge.

\paragraph{Causal contract for observed elevation.}
Elevation $z$ is used only as an optional relative quantity in factual relations, not as a target for full 3D dynamics. The model encodes $\operatorname{asinh}(\Delta z/1\,\mathrm{m})$ and a separate availability bit only when both endpoints of a relation are marked as observed by a trusted data converter. If elevation is missing at either endpoint, both the elevation value slot and the availability bit are set to zero, while the 2D relative geometry is retained. This distinguishes a genuine $\Delta z=0$ from missing elevation. Because closed-loop states generated by the model do not predict future $z$, their agent elevation is marked unobserved; neither future ground-truth elevation nor map-projected elevation may be reinjected into the factual encoder. Genuine observed relative elevation in the static map can nevertheless be preserved.

\subsection{Goal Regions and World-Conditioned Coordination}

\paragraph{Type-specific Goal proposals.}
For each agent, the model generates $K=32$ candidate Goal Regions over the 8-second horizon. Vehicles, pedestrians, and cyclists use separate type-specific prior banks. A pedestrian query is oriented using its most recent reliable direction of motion, falling back to the current heading at low speed. Define the element-wise smooth bound $\mathcal T_L(\bm z)=L\tanh(\bm z/L)$. The center and scale of candidate $k$ for agent $i$ are
\begin{align}
 \bm g^{q}_{i,k}&=\bm a_{i,k}+\mathcal T_{L_i}(\bm r_{i,k}),
 \qquad \bm g^s_{i,k}=\mathcal R_i\bm g^q_{i,k}+\bm p_i,\nonumber\\
 \bm s_{i,k}&=\operatorname{clip}\!\left(
 \bm s^0_{i,k}\odot\exp[0.75\tanh(\Delta\bm s_{i,k})],
 s_{\min},s_{\max}\right),
 \label{eq:goal-region}
\end{align}
where $q/s$ denote query/scene coordinates, $\bm a_{i,k}$ and $\bm s^0_{i,k}$ are type-specific priors, and $\mathcal R_i$ is defined by the current factual heading. Each proposal additionally carries map context, a coarse ETA interval, a unary score $u_{i,k}$, and a learned feasibility logit $f_{i,k}$. Its canonical score before World coordination is
\begin{equation}
 s_{i,k}=u_{i,k}+\alpha_{\mathrm{feas}}\sigma(f_{i,k}).
 \label{eq:proposal-score}
\end{equation}
This proposal stage reads only factual memory and therefore cannot exploit interaction outcomes that have not yet been generated.

\paragraph{World Queries.}
Eight learnable World Queries serve as scene-level seeds for future hypotheses. For scene $b$, the conditioning vector of World $w$ is
\begin{equation}
 \bm q_{b,w}=\operatorname{LN}\!\left(
 \bm e_w+\phi_s\!\left[\frac{1}{N_b}\sum_{i\in b}\bm h_i\right]\right),
 \label{eq:world-query}
\end{equation}
where $\bm h_i$ is the factual agent representation. Each World first adds a bounded conditional residual to the proposal scores and then performs two layers of sparse soft coordination over the factual agent graph. For World $w$ and agent $i$, the candidate distribution and its soft summary are
\begin{align}
 p_{w,i,k}&=\operatorname{softmax}_{k}(\ell_{w,i,k}),\qquad
 \bar{\bm e}_{w,i}=\sum_{k=1}^{K}p_{w,i,k}\bm e_{i,k},\nonumber\\
 \ell^0_{w,i,k}&=s_{i,k}+\Delta^0_{w,i,k}.
 \label{eq:world-soft-goal}
\end{align}
For each factual edge $j\!\rightarrow\! i$, the message jointly reads $(\bm h^w_j,\bm h^w_i,\bar{\bm e}_{w,j},\bar{\bm e}_{w,i})$, relative geometry, and the directed type pair, and is aggregated at the receiving agent through a learned gate. Each layer further produces a centered, bounded logit residual over valid proposals:
\begin{equation}
 \ell'_{w,i,k}=\ell_{w,i,k}+\Delta_{\max}
 \tanh\!\left(\delta_{w,i,k}-
 \frac{1}{|\mathcal K_i|}\sum_{k'\in\mathcal K_i}\delta_{w,i,k'}\right).
 \label{eq:world-coordination}
\end{equation}
This enables the Goal distribution of one agent to influence those of its neighbors within the same World without enumerating $32^N$ combinations. It is a tractable sparse-coupling approximation, not an exactly normalized global distribution. In this paper, a ``joint World'' specifically denotes a branch identity in which all agents share scene-level conditioning and are coupled through message passing. We do not claim to explicitly represent or normalize the full $32^N$ joint distribution, and we do not permit per-agent argmax selections to be stitched across Worlds.

After coordination, each World makes a single hard selection for every agent:
\begin{equation}
 k_{w,i}=\arg\max_k\ell_{w,i,k}.
\end{equation}
The Anchor class and its topological basin remain fixed throughout a standard 8-second rollout, preventing mode switches caused by commit-wise reselection. To permit continuous adaptation of the target as execution proceeds, we define the radial bound
\begin{equation}
 \mathcal B_R(\bm z)=
 \begin{cases}
 R\dfrac{\tanh(\|\bm z\|/R)}{\|\bm z\|}\bm z,&\|\bm z\|>0,\\
 \bm 0,&\text{otherwise},
 \end{cases}
 \label{eq:radial-bound}
\end{equation}
and, for $c>0$, update the center offset as
\begin{align}
 \bm d_{i,c}&=\mathcal R_{i,c}
 \mathcal B_{R_i^{\mathrm{step}}}(\Delta\bm o^q_{i,c}),\nonumber\\
 \bm o_{i,c}&=\mathcal B_{R_i^{\mathrm{total}}}
 (\bm o_{i,c-1}+\bm d_{i,c}),\qquad
 \widetilde{\bm g}_{i,c}=\bm g^s_{i,k_{w^*,i}}+\bm o_{i,c}.
 \label{eq:slow-goal-update}
\end{align}
The update reads the current facts, the persistent World condition, the remaining horizon, and the Preview endpoint and interaction summary from the previous commit. For any bounded scalar or vector $\bm x\in(a,b)$, define
\begin{align}
 \operatorname{Upd}_{[a,b]}(\bm x,\bm\delta;\eta)
 ={}&a+(b-a)\sigma\!\left[
 \operatorname{logit}\!\left(\frac{\bm x-a}{b-a}\right)\right.\nonumber\\
 &\left.\hspace{4.0em}+\eta\tanh(\bm\delta)\right],
 \label{eq:bounded-update}
\end{align}
which is used to update the Region scale and confidence separately. After several consecutive failed commits, a state is only marked as degraded or invalid; the discrete Anchor is not reselected. The Goal termination time is always fixed at the initial $t_0+8$ seconds, preventing recurrent planning from continually pushing the long-horizon target into the future.

\subsection{Preview-Induced Sparse Future Interaction}

For each World and each agent, the model generates four Preview nodes at $+0.5$, $+1.0$, $+1.5$, and $+2.0$ seconds. We refer to the two alternating rounds of Preview graph construction, interaction message passing, and plan refinement as the Joint Preview Interaction (JPI) block. Specifically, $\bm P_c^0$ induces $\mathcal E_c^0$, and $\operatorname{JPI}_1$ produces $\bm P_c^1$; the model then reconstructs $\mathcal E_c^1$ from $\bm P_c^1$, and $\operatorname{JPI}_2$ yields the final $\bm P_c^2$. JPI is an operator that updates the rolling plan, not a fourth persistent state, and it never writes directly to the factual history. Let $R_{i,c}$ be the remaining time to the Goal, $\bm v^q_{i,c}$ the current query-frame velocity, and $\bm d^q_{i,c}$ the query-frame displacement from the current position to the center of the Slow Goal. For Preview time $t_m$, we first construct a motion baseline with the correct terminal boundary condition:
\begin{align}
 \tau_{i,m}&=\min(t_m,R_{i,c}),\qquad
 s_{i,m}=\tau_{i,m}/R_{i,c},\nonumber\\
 h(s)&=s^2(3-2s),\nonumber\\
 \bm P^{\mathrm{base},q}_{i,m}
 &=\bm v^q_{i,c}\tau_{i,m}+h(s_{i,m})
 \left(\bm d^q_{i,c}-\bm v^q_{i,c}R_{i,c}\right).
 \label{eq:preview-baseline}
\end{align}
The initial nodes and the refinement at round $r$ are respectively
\begin{align}
 \bm P_{i,c}^{q,0}
 &=\bm P_{i,c}^{\mathrm{base},q}
 +\mathcal T_{L_P}(\bm r^0_{i,c}),\nonumber\\
 \bm P_{i,c}^{q,r+1}
 &=\bm P_{i,c}^{q,r}
 +\mathcal T_{0.35L_P}(\bm r^{r+1}_{i,c}).
 \label{eq:preview-residual}
\end{align}
Velocities are obtained by differencing adjacent 0.5-second nodes and are represented jointly with heading, progress, occupancy scale, and the nearest map topology. After 0.5 seconds are executed, the old $+1.0$, $+1.5$, and $+2.0$ second nodes are shifted to become the first three warm-start nodes of the next commit, while a new $+2.0$ second node is predicted. The warm start is used only as a conditioning feature and is never copied into the physical state.

\paragraph{Sparse future graph.}
For each directed agent pair $j\!\rightarrow\! i$, candidates are formed using the current envelope clearance $c^0_{ji}$, the minimum Preview-path clearance $c^f_{ji}$, closing speed, TTC, whether the current nearest polygons of the two agents coincide, and the geometric ETA difference $\Delta\eta^{\mathrm{geo}}_{ji}$. The ranking score is
\begin{align}
 \rho_{ji}={}&-\min(c^0_{ji},c^f_{ji})
 -0.25|\Delta\eta^{\mathrm{geo}}_{ji}|\nonumber\\
 &+\frac{2}{\min(\mathrm{TTC}_{ji},20)+1}
 +0.25\mathbb I_{\mathrm{same\mbox{-}map}}.
 \label{eq:edge-rank}
\end{align}
An edge is eligible whenever the current clearance lies within the local radius or the future clearance enters the warning range. Each receiving agent retains its highest-ranked neighbors, while edges whose current or future envelopes overlap are exempt from the neighbor cap. At round $r\in\{0,1\}$, the discrete edge set $\mathcal E_c^r$ is constructed from the current Preview $\bm P_c^r$, and continuous edge attributes are recomputed on that graph. Thus, the Preview modified by the first refinement can affect both the candidate topology and interaction strength in the second round. Discrete edge selection itself is not differentiated, whereas continuous geometry and message updates remain differentiable.

At round $r\in\{0,1\}$, edge attributes $\bm a^r_{ji}$ on $\mathcal E_c^r$ are used to predict the conflict logit $\kappa^r_{ji}$, signed arrival-time difference, and its scale:
\begin{align}
 \widehat{\Delta\eta}^{\,r}_{ji}
 &=\Delta\eta^{\mathrm{geo},r}_{ji}+1.5\tanh(\delta^r_{ji}),\nonumber\\
 \gamma^r_{ji}&=\sigma(\kappa^r_{ji}),\qquad
 b^r_{ji}=\exp\!\big(\operatorname{clip}(\beta^r_{ji},-4,2)\big),\nonumber\\
 p^r(j\prec i)&=\sigma\!\left(-\widehat{\Delta\eta}^{\,r}_{ji}/b^r_{ji}\right).
 \label{eq:interaction-timing}
\end{align}
The conflict probability $\gamma^r_{ji}$ also gates the sparse message from $j$ to $i$. Aggregated messages update the agent-level World hidden state and refine the four Preview nodes:
\begin{equation}
 (\bm P_c^{r+1},\bm h_{c}^{w,r+1})=
 \operatorname{JPI}_{r+1}(\bm P_c^r,\bm h_c^{w,r},\bm F_c,
 \widetilde{\bm g}_c;\mathcal E_c^r,\bm a_c^r,\bm\gamma_c^r).
 \label{eq:preview-refinement}
\end{equation}
Precedence is derived monotonically from signed $\Delta$ETA rather than predicted by an unconstrained priority head that could contradict the timing difference. Importantly, the two JPI rounds and their intermediate graph reconstruction update only the rolling Preview, interaction structure, and World hidden state; they do not alter the fixed Goal Anchor. The Slow Goal is updated only at the beginning of the next commit, where it uses the interaction summary left by the current commit in Eq.~\eqref{eq:slow-goal-update}. Ground-truth positive conflict edges added during training are used solely to compute interaction losses and never enter inference-time message passing, thereby preventing future-topology leakage.

\subsection{Causal World Plan Scoring}

At $c=0$, after all eight Worlds have completed lightweight Goal and Preview planning, the WorldPlanScorer predicts a six-dimensional quality vector $\widehat{\bm q}_{b,w}$ and an aggregate score $s_{b,w}$ for every scene--World pair. The six components correspond to Goal, Preview, Interaction, Map, Dynamics, and Closed-loop quality. Let $\bm z_{b,w}$ aggregate the Preview and World representations, Goal geometry, interaction timing, and committed quality prefix of that World. Then
\begin{equation}
 \widehat{\bm q}_{b,w}=\psi_q(\operatorname{sg}[\bm z_{b,w}]),
 \qquad
 s_{b,w}=\psi_s(\operatorname{sg}[\bm z_{b,w}],\widehat{\bm q}_{b,w}),
 \label{eq:world-scorer}
\end{equation}
where $\operatorname{sg}$ stops planning gradients from the Scorer. Map, safety, dynamics, and trajectory losses still supervise the generator directly, preventing the planner from altering its plan features merely to please the internal Scorer. At inference time, the model selects $w^*=\arg\max_w s_{b,w}$ and retains that World identity for all 16 commits of the same scene. Subsequent causal scores evaluate only the selected branch and never stitch together agent-wise Worlds.

Supervised training likewise executes only one physical World per scene. Scene-level WTA routing, the causal Scorer winner, and balanced exploration are mixed to select the branch. For a given scene, only the routed World receives full physical-execution supervision; routing coverage over training gives different Worlds opportunities to receive such labels. WTA selects a branch but never injects a ground-truth Anchor into the forward Goal process. Unexecuted Worlds receive only the same lightweight plan-level supervision, and a two-second proxy cannot stand in for realized eight-second closed-loop quality. Scorer supervision combines the initial lightweight quality over eight Worlds, the per-commit causal quality of the selected branch, and its accumulated realized quality:
\begin{equation}
 \mathcal L_{\mathrm{score}}
 =\beta_{\mathrm{light}}\mathcal L_{\mathrm{light}}^{1:W}
 +\mathcal L_{\mathrm{causal}}^{w^*}
 +\beta_{\mathrm{real}}\mathcal L_{\mathrm{realized}}^{w^*}.
 \label{eq:scorer-loss}
\end{equation}
The final term uses stop-gradient initial plans and accumulated realized quality to calibrate whether the initial plan predicts long-horizon execution quality. Since only $w^*$ receives a complete label, this term must not be interpreted as genuine full-ranking supervision over all eight Worlds.

\subsection{Continuous Control and Differentiable Dynamics}

The control head reads the final Preview, Slow Goal, World state, factual memory, and current physical state, and predicts a continuous distribution over the next second. For each agent, the latent variable of the flattened 30-dimensional control sequence is
\begin{align}
 \bm z={}&\bm\mu+\exp(\log\bm\sigma)\odot\bm\epsilon
 +\bm L\bm\xi,\nonumber\\
 &\bm\epsilon\sim\mathcal N(0,\bm I),\qquad
 \bm\xi\sim\mathcal N(0,\bm I_2),
 \label{eq:control-distribution}
\end{align}
where $\bm L$ is a rank-2 temporal covariance factor. Stochastic export further uses temporally correlated noise,
\begin{equation}
 \bm\epsilon_t=\rho\bm\epsilon_{t-1}
 +\sqrt{1-\rho^2}\,\bm\eta_t,\qquad
 \bm\eta_t\sim\mathcal N(\bm 0,\bm I),
 \label{eq:ar-noise}
\end{equation}
rather than independent white noise at every step. At each 0.1-second step, the 2D query-frame acceleration is mapped onto a disk, while the yaw acceleration is mapped to an interval:
\begin{equation}
 \bm a_t^q=A_i\frac{\tanh(\|\bm z^{xy}_t\|)}{\|\bm z^{xy}_t\|}\bm z^{xy}_t,
 \qquad
 \alpha_t=\Omega_i\tanh(z^{\omega}_t).
 \label{eq:control-squash}
\end{equation}
The continuous limit is used as $\|\bm z^{xy}_t\|\rightarrow0$. Vehicles and cyclists retain all three channels, whereas the yaw-control channel is masked for pedestrians, which use only 2D holonomic acceleration. The invertible transformations and their Jacobians are included in the control NLL so that bounded physical controls are not incorrectly modeled as unconstrained Gaussian variables.

After rotating query-frame controls into the scene frame, all agents are integrated synchronously in FP32:
\begin{align}
 \bm v_{t+1}&=\bm v_t+\bm a_t^{s}\Delta t,\\
 \bm p_{t+1}&=\bm p_t+\tfrac{1}{2}(\bm v_t+\bm v_{t+1})\Delta t,\nonumber\\
 \omega_{t+1}&=\operatorname{clip}(\omega_t+\alpha_t\Delta t),\nonumber\\
 \theta_{t+1}&=\operatorname{wrap}\!\left(
 \theta_t+\tfrac{1}{2}(\omega_t+\omega_{t+1})\Delta t\right).
 \label{eq:dynamics}
\end{align}
Vehicles and cyclists explicitly maintain yaw and yaw rate. A pedestrian's heading is determined by its direction of motion when its speed is sufficiently high and otherwise retains the previous heading. Extreme clipping in the dynamics is used only as an emergency numerical safeguard and is not responsible for learned collision avoidance. Preview nodes and local ground-truth states supervise the Preview, while the trajectory obtained by integrating the mean controls is supervised by continuous ground-truth states. Because Control is conditioned on the final Preview representation, state, safety, and map losses can backpropagate to the Preview through the control path; however, the model does not introduce an additional direct alignment objective between the four two-second Preview nodes and the one-second control trajectory. At inference time, Control does not overwrite Preview coordinates, and physical states are produced only by Eq.~\eqref{eq:dynamics}.

\subsection{Learning Objectives and Streaming Generated-State Training}

The learning objectives are grouped by the interfaces they supervise: Goal proposal and Region; World assignment and diversity; Slow Goal region and continuity; Preview state, continuity, and topology; interaction conflict, timing, and reciprocity; World Scorer; control distribution; integrated state; kinematic priors; safety; map feasibility; and closed-loop error. To prevent the number of vehicles from overwhelming rare agent types, agent-level terms are first averaged over time and agents and then macro-averaged across agent types present in the current batch. Interaction terms are macro-averaged across supervised directed type pairs, whereas global safety terms retain the within-scene pair average. In general, let $\mathcal G_m^+$ be the set of reduction groups containing valid elements for loss $m$. We compute
\begin{align}
 \overline{\ell}_{m,g}
 &=\frac{\sum_{j\in\mathcal V_{m,g}}\ell_{m,j}}
 {\max(1,|\mathcal V_{m,g}|)},\nonumber\\
 \overline{\mathcal L}_m
 &=\frac{1}{\max(1,|\mathcal G_m^+|)}
 \sum_{g\in\mathcal G_m^+}
 \overline{\ell}_{m,g},\nonumber\\
 \mathcal L
 &=\sum_m\lambda_m(c)\,
 \frac{\overline{\mathcal L}_m}
 {\max\{1,\operatorname{EMA}(|\overline{\mathcal L}_m|)\}}.
 \label{eq:loss}
\end{align}
Terms with empty support produce no gradient. Future interaction geometry, dynamics, key probability normalizations, and loss denominators are evaluated in FP32. The non-amplifying EMA balances scale differences without magnifying auxiliary terms that are intrinsically small. Kinematic terms supervise ground-truth acceleration and regularize jerk, sideslip, and lateral acceleration; safety and map terms operate on scene-level joint trajectories after dynamics integration. Missing future ground truth, missing elevation, and invalid interaction timing are controlled by their respective masks.

Unlike training exclusively on logged states, the training rollout writes the first five states generated by Eq.~\eqref{eq:closed-loop-recursion} into the history of the next commit. Our evaluation uses the step-120k Stage~1 checkpoint learned from the first two commits starting from a real 11-frame history. Streaming training partitions the $C$ commits into finite TBPTT windows. Numerical state evolves continuously from model-generated commits, while $\operatorname{sg}$ at a window boundary truncates only the computation graph; it neither reinjects ground truth nor resets the physical state. TBPTT limits memory usage and the temporal extent of gradient credit assignment, while preserving continuity of generated states across windows.

\subsection{Prefix-Frozen Generated-State Recovery}

When a single shared model is optimized only through later commits, gradients still update the same parameters that generate the early prefix. Consequently, the input distribution faced by the recovering suffix changes during optimization, and suffix gradients can degrade already learned prefix behavior. S2.1 introduces a prefix-frozen temporal cascade that aligns temporal responsibilities with parameter ownership. Models A and B have identical architectures but disjoint parameter sets, both initialized from S1 step~120k. A remains permanently in evaluation mode and is excluded from the optimizer, scheduler, and AMP scaler; B is the only model updated. With zero-based global commit sets and their executed time intervals defined as
\begin{align}
 \mathcal C_A&=\{0,1\},\qquad
 \mathcal C_B=\{2,3\},\nonumber\\
 \tau(c)&=[0.5c,0.5(c+1))\ \mathrm{s},
 \label{eq:temporal-cascade-time}
\end{align}
A executes the first two commits from the real 11-frame history to produce a model-generated prefix over 0--1 seconds. B neither executes nor optimizes C1/C2; it starts from the handoff state at $t=1$ second and learns recovery over 1--2 seconds.

The boundary is not a hidden-state distillation interface, but an auditable typed causal interface. Let $\bm z^{\mathrm{phys}}_1$ contain the latest 11-frame generated history, current kinematic state, active/lifecycle state, map context, and causally available signal context; let $w_A$ be the integer World index selected by A; and let $\bm\mu$ encode scene identity and generated-history provenance. Then
\begin{align}
 \bm s_1^A
 &=\operatorname{Rollout}_{\theta_A}(\bm H_0;\mathcal C_A),\nonumber\\
 \bm z^{A\rightarrow B}_1
 &=\operatorname{sg}\!\left(
   \bm z^{\mathrm{phys}}_1(\bm s_1^A),\,w_A,\,\bm\mu
 \right),\nonumber\\
 \left(\bm F_1^B,\{\bm b_1^{B,w}\}_{w=1}^{8}\right)
 &=\operatorname{Rebuild}_{\theta_B}
   \!\left(\bm z^{\mathrm{phys}}_1\right),
 \qquad w_B\leftarrow w_A.
 \label{eq:temporal-cascade-handoff}
\end{align}
All tensors passed through the boundary are detached and cloned. Encoder hidden states, KV/cache, decoder memory, Slow Goal, Goal/World features, interaction summaries, Preview warm starts, and the computation graph of A are discarded. B reruns the factual encoder, regenerates Goal proposals and World representations as well as its own Slow Goal, and reconstructs interaction from the new Preview and physical state. The active state, causal-exit state, and its confirmation count are carried forward as explicit lifecycle facts. Equation~\eqref{eq:temporal-cascade-handoff} preserves routing through the same discrete World index but neither transfers nor presupposes semantic continuity between the neural hidden states of A and B.

S2.1 optimizes B only on the generated-state continuation at global Commits~3/4:
\begin{align}
 \mathcal L_{\mathrm{S2.1}}(\theta_B;\theta_A)
 &=\sum_{c=2}^{3}\sum_{m\in\mathcal M_{\mathrm{rec}}}
 \lambda_m\overline{\mathcal L}_{m,c},\nonumber\\
 \mathcal L^{B}_{0}&=\mathcal L^{B}_{1}=0,\qquad
 \theta_A^{k+1}=\theta_A^k,\nonumber\\
 \theta_B^{k+1}
 &=\theta_B^k-\eta\nabla_{\theta_B}
 \mathcal L_{\mathrm{S2.1}}.
 \label{eq:temporal-cascade-loss}
\end{align}
Here, $\mathcal M_{\mathrm{rec}}$ includes Slow Goal continuity as well as the recursive Preview, Interaction, Control, State, Closed-loop, Safety, Map, and Kinematic objectives; static Goal proposal, World Scorer, and World diversity terms are disabled. Supervision for the two B commits begins at future steps 10 and 15, respectively. During training, a detached GT-WTA proxy that depends only on future indices 14/19 may select the branch and supply supervised edges for C3/C4. Perturbing future$[0{:}10]$ does not change the winner, and ground truth is never written into generated history. During target-free validation and deployment, A selects $w_A$ using its causal Scorer and, after re-encoding, B is routed only along the same index.

\FloatBarrier

%% file: figures/hi_floop_architecture.tex
\begin{figure*}[t]
  \centering
  \includegraphics[width=\linewidth]{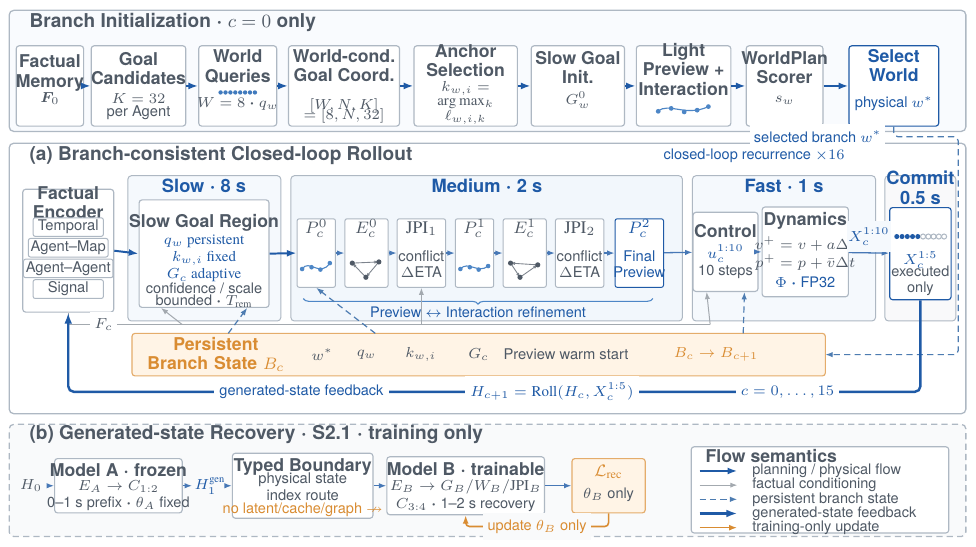}
  \caption{Architecture of \model{}. At initialization, eight scene-level Worlds propose joint future branches and the scorer selects one physical World whose identity persists through all 16 commits. The recurrent core updates an 8-s Goal, alternates two rounds of 2-s Preview--Interaction refinement, decodes 1-s Control, and writes only the executed 0.5-s prefix back to factual history. The S2.1 path uses a frozen A$\rightarrow$B recovery cascade: its typed boundary transfers physical state and causal context, but no latent state, cache, or gradient graph.}
  \label{fig:architecture}
\end{figure*}

%% file: sections/experiments_conclusion.tex
\section{Experiments}
\label{sec:experiments}

We evaluate the completed S2.1 A$\rightarrow$B cascade on the full H-D public-validation split of 955 scenarios using an eight-second closed-loop rollout. Frozen Model~A is initialized from S1 step~120k and executes Commit~1--2; Model~B at step~100k re-encodes the physical handoff and executes Commit~3--16. Scenario-joint displacement is the primary result because it measures whether one World explains all evaluated agents in a scene, while per-agent World selection is retained as a complementary diagnostic of marginal mode coverage.

\subsection{Datasets and Implementation Details}

Five datasets are converted into a unified ragged scene representation through dataset-specific adapters~\cite{ettinger2021womd,wilson2021argoverse2,chen2026hetrod,bock2020ind,levelx2021unid}. The step-120k S1 checkpoint uses fixed Waymo/AV2/H-D training splits, whereas S2.1 uses a five-source, train-only manifest. Table~\ref{tab:data-role} summarizes the scheduled source allocation over 100k optimization steps and the corresponding held-out boundaries. Approximately $3.2\%$ of the Waymo scenarios that do not strictly follow the 10\,Hz sampling rate are removed according to the predefined filtering rule. Future traffic signals are supplied exogenously by the environment when available and otherwise fully masked. Missing elevation is likewise strictly masked; the model does not impute either elevation or future signal states.

\begin{table*}[!b]
\centering
\caption{Five-source train-only sampling contract for S2.1 and the H-D evaluation split. The allocation column reports the planned source allocation accumulated over 100k optimization steps.}
\label{tab:data-role}
\footnotesize
\begin{tabularx}{\textwidth}{P{2.8cm}P{2.7cm}rY}
\toprule
Dataset & Training identity & 100k allocation & Evaluation role\\
\midrule
H-D & Official training split, S1-matched release & 48.0k & Public-valid 955 as an independent evaluation set\\
WOMD & Official training split & 19.2k & Official validation split as a held-out evaluation set\\
Argoverse~2 & Official training split & 11.0k & Official validation split as a held-out evaluation set\\
inD & Train only & 13.6k & Valid/audit/test remain held out\\
uniD & Train only & 8.2k & Valid/audit/test remain held out\\
H-D public-valid & Evaluation only, 955 scenarios & -- & Eight-second closed-loop evaluation cohort\\
\bottomrule
\end{tabularx}
\end{table*}

Throughout S2.1, source weights are reported in the fixed order H-D/WOMD/AV2/inD/uniD. The respective weights are $35\%$, $25\%$, $15\%$, $15\%$, and $10\%$ during steps $0$--$40$k; $50\%$, $18\%$, $10\%$, $14\%$, and $8\%$ during steps $40$--$80$k; and $70\%$, $10\%$, $5\%$, $10\%$, and $5\%$ during steps $80$--$100$k. The manifest contains no validation entries, and the training entry point fails closed if any path or name contains \texttt{valid}, \texttt{validation}, \texttt{audit}, or \texttt{test}.

The main model uses an embedding dimension of $d=384$, 32 Goal Regions, eight Worlds, four Preview nodes, and two rounds of JPI. The S1 instance contains 114,106,461 trainable parameters. Learnable modules operate in BF16, whereas future-interaction geometry, dynamics, key probabilistic operations, and loss reductions are evaluated in FP32. The probability-domain BCE for the Slow Goal is computed in an autocast-disabled FP32 region without changing the mathematical objective or gradient definition. S2.1 was trained for 100k optimization steps with a physical batch size of 16 on a single 80-GB H100. Models A and B are two complete, independent model instances, and the optimizer contains only the parameters of B.

\subsection{Evaluation Protocol and Metrics}

We perform a full eight-second closed-loop rollout on all 955 scenarios in the H-D public-validation split; all scenarios complete successfully, with no skipped or failed cases. The model produces eight learned Worlds and four temporally persistent Control samples per World, giving the 32 stochastic rollouts required by the official evaluator. Probabilities used by the World-level Brier metrics are defined over the eight learned Worlds; the four within-World Control samples are not treated as additional semantic modes. Throughout this section, $K$ denotes the number of candidate Worlds or rollouts, not the number of agents; the suffix @8 denotes eight learned Worlds rather than the temporal horizon.

For scenario $s$, horizon $\tau$, evaluated-agent set $\mathcal A_s(\tau)$, and World $w$, let $\operatorname{ADE}_{s,i,w}(\tau)$ and $\operatorname{FDE}_{s,i,w}(\tau)$ denote the displacement errors of agent $i$. Scenario-joint evaluation selects one World for the entire scene,
\begin{equation}
 w_s^\star(\tau)=\arg\min_{1\leq w\leq 8}
 \frac{1}{|\mathcal A_s(\tau)|}
 \sum_{i\in\mathcal A_s(\tau)}\operatorname{FDE}_{s,i,w}(\tau),
 \label{eq:scenario-joint-selection}
\end{equation}
and reports both ADE and FDE from this same $w_s^\star(\tau)$, denoted ADE-at-joint-minFDE@8 and joint-minFDE@8, respectively. Thus, every agent in a scene is evaluated under a shared joint-future hypothesis. For the secondary agent-centric diagnostic, each agent instead selects
$w_{s,i}^\dagger=\arg\min_w\operatorname{FDE}_{s,i,w}$; we report FDE at this mode and ADE from the same mode, denoted ADE-at-minFDE. We additionally define $\widehat w_{s,i}=\arg\min_w\operatorname{ADE}_{s,i,w}$ and report the independently ADE-selected oracle-minADE. Neither diagnostic is a jointly realizable World because different agents may select different modes.

\FloatBarrier
\subsection{Full-955 Closed-Loop Results}
\label{sec:full955-results}

\paragraph{Scenario-joint displacement}
Table~\ref{tab:scenario-joint-results} evaluates the eight learned Worlds under the shared selection rule in Eq.~\eqref{eq:scenario-joint-selection}. The short-horizon values quantify accurate local continuation, while the complete eight-second rollout reaches a scenario-joint ADE-at-joint-minFDE@8/joint-minFDE@8 of $2.047953/6.383503$\,m. Crucially, each number is obtained from one World selected for all evaluated agents in a scene rather than from a post-hoc composition of agent-wise modes.

\begin{table}[!htbp]
\centering
\caption{Scenario-joint displacement over the full H-D 955-scenario evaluation. At every horizon, a single selected World explains all evaluated agents in the scene.}
\label{tab:scenario-joint-results}
\small
\begin{tabular}{ccc}
\toprule
Horizon & \shortstack{ADE-at-joint-\\minFDE@8 (m)} & \shortstack{joint-minFDE@8\\(m)}\\
\midrule
1 s & 0.051287 & 0.067154\\
2 s & 0.126420 & 0.322037\\
3 s & 0.272011 & 0.771361\\
4 s & 0.481560 & 1.386835\\
5 s & 0.755734 & 2.211627\\
6 s & 1.101594 & 3.301054\\
7 s & 1.530270 & 4.685808\\
8 s & 2.047953 & 6.383503\\
\bottomrule
\end{tabular}
\end{table}

\paragraph{Cross-benchmark context}
To place the displacement scale in context, Table~\ref{tab:cross-benchmark-context} juxtaposes public open-loop motion-forecasting results with the H-D closed-loop diagnostics above while retaining each benchmark's native protocol. The \model{} agent-centric rows use the same $K=8$ selection budget at both six and eight seconds. The additional eight-second Scene $K=8$ row instead requires one World to explain all evaluated agents jointly.

% Public rows: cited papers and the corrected NDPNet repository result.
% Hi-FLoop rows: frozen S2.1 step-100k full-955 all-horizon evidence under
% evidence/s21_ab_step100k_all_horizon_20260908_v1/.
\begin{table*}[!t]
\centering
\caption{Contextual displacement reference under native benchmark protocols. Published rows use their original test protocols; \model{} uses the H-D public-valid full-955 closed-loop protocol. Published and agent-centric rows report independently selected minADE and minFDE, whereas the scene-joint row reports ADE-at-joint-minFDE and joint-minFDE from one shared World. Values are not directly comparable across datasets, horizons, aggregation schemes, or selection rules. The NDPNet WOMD entry follows the corrected epoch-31 result released with the \href{https://github.com/HuaHu-yizhou/NDPNet}{official implementation}.}
\label{tab:cross-benchmark-context}
\footnotesize
\setlength{\tabcolsep}{5pt}
\renewcommand{\arraystretch}{1.02}
\begin{tabularx}{\textwidth}{Y P{1.35cm} P{2.25cm} P{2.05cm} r r}
\toprule
Method & Dataset & Horizon & Selection & ADE (m) & FDE (m)\\
\midrule
NDPNet~\cite{hu2026ndpnet} & WOMD & 8 s & Agent $K=6$ & 0.840 & 1.751\\
NDPNet~\cite{hu2026ndpnet} & AV2 & 6 s & Agent $K=6$ & 0.610 & 1.170\\
MTR++ ensemble~\cite{shi2023mtrpp} & WOMD & 3/5/8 s avg. & Agent $K=6$ & 0.558 & 1.117\\
MTR++~\cite{shi2023mtrpp} & WOMD & 3/5/8 s avg. & Agent $K=6$ & 0.591 & 1.194\\
QCNet ensemble~\cite{zhou2023qcnet} & AV2 & 6 s & Agent $K=6$ & 0.620 & 1.190\\
QCNet~\cite{zhou2023qcnet} & AV2 & 6 s & Agent $K=6$ & 0.650 & 1.290\\
MTR~\cite{shi2022mtr,zhou2023qcnet} & AV2 & 6 s & Agent $K=6$ & 0.730 & 1.440\\
\midrule
\textbf{Hi-FLoop (ours)} & H-D & 6 s & Agent $K=8$ & 0.526 & 1.621\\
\textbf{Hi-FLoop (ours)} & H-D & 8 s & Agent $K=8$ & 0.875 & 2.810\\
\textbf{Hi-FLoop (ours, joint)} & H-D & 8 s & Scene $K=8$ & 2.048 & 6.384\\
\bottomrule
\end{tabularx}
\end{table*}

\paragraph{Agent-centric diagnostic}
Table~\ref{tab:agent-centric-oracle} reports the agent-centric oracle displacement across evaluation horizons. When each agent independently selects its ADE-optimal World among all eight modes, oracle-minADE@8 is $0.526000$\,m at six seconds and $0.874687$\,m at eight seconds; independent FDE-based selection gives minFDE@8 of $1.621078$\,m and $2.809677$\,m, respectively. These lower agent-centric errors diagnose marginal mode coverage; because the selected World may differ across agents and between the two metrics, they do not replace the scenario-joint results above.

\begin{table}[!t]
\centering
\caption{Agent-centric oracle displacement at every integer-second horizon on H-D full-955. Each column independently selects its best World per agent; only agents with a valid endpoint at that horizon are included.}
\label{tab:agent-centric-oracle}
\footnotesize
\setlength{\tabcolsep}{4pt}
\renewcommand{\arraystretch}{0.96}
\begin{tabular}{ccc}
\toprule
Horizon & \shortstack{oracle-minADE@8\\(m)} & \shortstack{minFDE@8\\(m)}\\
\midrule
1 s & 0.034846 & 0.049019\\
2 s & 0.080559 & 0.191897\\
3 s & 0.153480 & 0.406649\\
4 s & 0.250543 & 0.706843\\
5 s & 0.374106 & 1.108023\\
6 s & 0.526000 & 1.621078\\
7 s & 0.694877 & 2.198541\\
8 s & 0.874687 & 2.809677\\
\bottomrule
\end{tabular}
\end{table}

\subsection{Analytical Mechanism Decomposition}
\label{sec:structural-checks}

The complete \model{} contract enforces the scene-level branch identity $w_c=w_0$, current-fact encoding $\bm F_c=E(\bm H_c,\bm M)$, and the executed-prefix transition $\bm H_{c+1}=\operatorname{Roll}(\bm H_c,\bm X_c^{1:5})$. Table~\ref{tab:structural-checks} analytically decomposes the framework by changing one contract condition at a time and tracing the structural property that is consequently removed.

\begin{table*}[!t]
\centering
\caption{Analytical decomposition of the mechanisms in \model{}. Each row changes one contract condition and identifies the structural property that is consequently removed.}
\label{tab:structural-checks}
\footnotesize
\begin{tabularx}{\textwidth}{P{2.7cm}P{5.1cm}Y}
\toprule
Mechanism check & Controlled alteration & Consequence under the model definition\\
\midrule
Persistent joint branch & Reselect $w_c=\arg\max_w s_{c,w}$ at every commit & Adjacent commits may follow different long-horizon hypotheses, so temporal branch consistency is no longer guaranteed\\
World semantics & Replace eight Worlds by one World with 32 Control samples & Execution stochasticity remains, but scene-level diversity of joint hypotheses is removed; sample count alone does not recover multi-World semantics\\
Executed-state feedback & Hold the factual representation fixed at $E(\bm H_0,\bm M)$ while dynamics advances & Subsequent Goals, Previews, and Controls cannot condition on generated changes in position, neighborhood, or conflict timing\\
Multi-timescale state & Remove the persistent Goal and two-second Preview and decode one-second Control directly & No separate state carries long-term behavioral direction or a mid-term interaction plan, collapsing the explicit temporal decomposition\\
Future-graph message & Zero the JPI graph-message residual and remove conflict/$\Delta$ETA conditioning from Control & Agents remain coupled by factual attention and the shared World, but predicted future conflicts and passing order cease to condition local motion\\
\bottomrule
\end{tabularx}
\end{table*}

Together, these checks isolate branch persistence, executed-state feedback, temporal decomposition, and explicit future-interaction conditioning as distinct properties of the proposed computation contract.

\subsection{S1 Optimization Diagnostic}
\label{sec:s1-optimization}

\begin{figure}[!htbp]
\centering
\includegraphics[width=\linewidth]{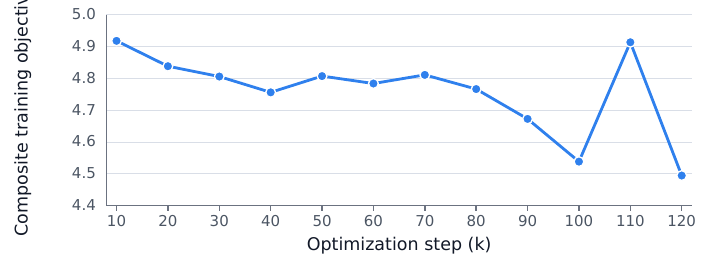}
\caption{S1 optimization diagnostic. The curve reports the composite training objective $\mathcal L$ in Eq.~\eqref{eq:loss} at 10k-step checkpoints during the 120k-step run; markers denote logged values, and the connecting line is not smoothed.}
\label{fig:s1-total-loss}
\end{figure}

Figure~\ref{fig:s1-total-loss} records the composite training objective at 10k-step intervals for the S1 run used to initialize frozen Model~A. We include this curve solely as an optimization diagnostic; its values are training objectives and are not used as evidence of held-out or closed-loop performance.

\FloatBarrier
\section{Conclusion}

We presented \model{}, a branch-consistent, multi-timescale state-feedback framework for closed-loop generation. Within a fixed scene-level joint World, the long-horizon Goal, rolling JPI plan, and short-horizon Control operate with distinct horizons, temporal persistence, and update semantics. At each commit, only the dynamically executed 0.5-s prefix is appended to the history and re-encoded as factual context for the next iteration. The scene-level World preserves a shared branch identity across agents and commits, while the Goal, interaction relations, and local motion adapt to newly generated facts within that branch. The A$\rightarrow$B temporal cascade extends this principle to generated-state recovery through index-consistent routing and latent-state reconstruction across a typed physical boundary, while disjoint parameters isolate suffix gradients from the frozen prefix model. On all 955 H-D public-validation scenarios, the S2.1 cascade obtains an eight-second ADE-at-joint-minFDE@8/joint-minFDE@8 of $2.048/6.384$\,m when one World must jointly explain every evaluated agent; independent per-agent ADE selection gives oracle-minADE@8 of $0.526$\,m at six seconds and $0.875$\,m at eight seconds.

\section*{Acknowledgments}

We thank Professor Zhan-Gang Han of the School of Systems Science, Beijing Normal University, for his guidance and support. We also gratefully acknowledge the support of the School of Systems Science, Beijing Normal University. The unified data interface in this work covers the Waymo Open Motion Dataset, Argoverse~2, and H-D (HetroD), while the S2.1 train-only mixture additionally uses inD and uniD to complement heterogeneous road-user and shared-space traffic scenarios. We thank the respective dataset maintainers for making these resources available~\cite{ettinger2021womd,wilson2021argoverse2,chen2026hetrod,bock2020ind,levelx2021unid}. We further acknowledge Joint Metrics Matter and the Waymo Open Sim Agents Challenge for advancing the evaluation of joint futures and closed-loop realism~\cite{weng2023jointmetrics,montali2023wosac}, as well as Waymax and ScenarioNet for providing open infrastructure that accelerates simulation and cross-dataset scene management~\cite{gulino2023waymax,li2023scenarionet}. The precise data splits, adaptation rules, and scope of tool usage follow the experimental setup described above.